\documentclass[a4paper,11pt]{article}

\usepackage{fourier}
\usepackage{color}
 \usepackage{graphicx}
 \usepackage{subfig}
\usepackage{url}
\usepackage[affil-it]{authblk}
\usepackage{amsmath}
\usepackage{wrapfig}

\usepackage[T1]{fontenc}
\usepackage{times}
\usepackage{gensymb}

\begin{document}

\title{A lightweight 3D dense facial landmark estimation model from position map data}

\author[1,2,*]{Shubhajit Basak}
\author[2]{Sathish Mangapuram}
\author[2]{Gabriel Costache}
\author[3]{Rachel	McDonnell}
\author[1]{Michael Schukat}
\affil[1]{School Of Computer Science, University of Galway}
\affil[2]{Xperi Corporation}
\affil[3]{School of Computer Science and Statistics, Trinity College Dublin}
\date{}
\maketitle
\thispagestyle{empty}

\begin{abstract}
The incorporation of 3D data in facial analysis tasks has gained popularity in recent years. Though it provides a more accurate and detailed representation of the human face, accruing 3D face data is more complex and expensive than 2D face images. Either one has to rely on expensive 3D scanners or depth sensors which are prone to noise. An alternative option is the reconstruction of 3D faces from uncalibrated 2D images in an unsupervised way without any ground truth 3D data. However, such approaches are computationally expensive and the learned model size is not suitable for mobile or other edge device applications. Predicting dense 3D landmarks over the whole face can overcome this issue. As there is no public dataset available containing dense landmarks, we propose a pipeline to create a dense keypoint training dataset containing 520 key points across the whole face from an existing facial position map data. We train a lightweight MobileNet-based regressor model with the generated data. As we do not have access to any evaluation dataset with dense landmarks in it we evaluate our model against the 68 keypoint detection task. Experimental results show that our trained model outperforms many of the existing methods in spite of its lower model size and minimal computational cost. Also, the qualitative evaluation shows the efficiency of our trained models in extreme head pose angles as well as other facial variations and occlusions. Code is available at: https://github.com/shubhajitbasak/dense3DFaceLandmarks
\end{abstract}

\section{Introduction}
Predicting the 3D features of the human face is the pre-requisite for many facial analysis tasks such as face reenactment and speech-driven animation, video dubbing, projection mapping, face replacement, facial animations, and many others \cite{zollhofer2018state}. Due to the limitation of depth sensors, it is difficult to capture high-frequency details through RGB-D data. Capturing high-quality 3D scans is expensive and often restricted because of ethical and privacy concerns. A popular alternative to these facial capturing methods is to estimate the face geometry from an uncalibrated 2D face image. However, this 3D-from-2D reconstruction of the human face is an ill-posed problem because of the complexity and variations of the human face. With the help of highly complex deep neural network models, we are able to recover the detailed face shape from uncalibrated face images. However, most of these methods depend on some kind of statistical priors of face shape like a 3D morphable model(3DMM) and the sparse face landmarks for face alignments. Some of the previous works also used additional signals beyond color images, like  facial depth \cite{ bao2021high}, optical flow \cite{cao2018stabilized}, or multi-view stereo \cite{beeler2010high}, and then optimized their methods using geometric and temporal prior. Each of these methods can produce very detailed results, but take a very long time to compute. At the same time, the model size and the huge computational requirements make these approaches not suitable for real-time applications in edge devices. Therefore it is still a very challenging task to implement a face modeling pipeline on limited computational cost systems such as mobile or embedded devices.

Estimating dense 3D landmarks on the face through facial landmark detection (FLD) can work as an alternative to estimating the face structure. The goal of FLD is to localize predefined feature points on the 2D human face such as the nose tip, mouth, eye corners, etc., which have anatomical importance. Almost all of the FLD tasks try to predict sparse landmarks on the face which comprises 68 key points both in 2D or 3D space. Specifically due to its robustness to illumination and pose variations, 3D FLD task has gained increasing attention among the computer vision community. Unfortunately, this set of sparse landmarks fails to encode most of the intricate facial features. So increasing the number of these landmarks can help to learn face geometry better. However, publicly available datasets mostly contain a sparse set of 68 facial landmarks, which fails to encode the full face structure. To achieve a high landmark density we take the existing approach of \cite{feng2018joint} which produces a position map data of the face. We propose a sampling methodology to filter 520 key points from the whole face and create a dataset to train a lightweight regression model. As we do not have access to any public dataset which has dense landmarks, we evaluated the trained regression model against the 68 landmark estimation task. Experimental result shows that the trained regression model can produce comparable result in the 3D face alignment task.

\section{Related Work}

Annotating a real face with dense landmarks is highly ambiguous and expensive. Some of the previous methods, like Wood et al.\cite{wood2021fake}, rely on synthetic data alone. Though the authors have detailed ground truth annotations like albedo, normals, depth, and dense landmarks, none of these data is publicly available. The authors also proposed a method \cite{wood20223d} to learn the dense landmarks as a Gaussian uncertainty from those synthetic data and fit a 3DMM model from those dense key points only. Some other methods \cite{deng2020retinaface, feng2018joint} use pseudo-labels model-fitting approaches like fitting an existing 3DMM model to generate synthetic landmarks. \cite{jeni2015dense} predicted dense frontal face landmarks with cascade regressions. Through an iterative method, they created 1024 dense 3D landmark annotations from 3D scan datasets \cite{ zhang2014bp4d}. In contrast, Kartynnik et al. \cite{kartynnik2019real} used a predefined mesh topology of 468 points arranged in fixed quads and fit a 3DMM model to a large set of in-the-wild images to create ground truth 3D dense annotations of key points. They later employed direct regression to predict these landmarks from face images. Some other methods \cite{alp2017densereg, feng2018joint} used a different method to unwrap each pixel of the face as a position map and regress the position in 3D space. They created the position map by fitting the Basel Face Model (BFM) \cite{paysan20093d} from the 300WLP dataset \cite{zhu2016face}, which has the 3DMM parameters associated with more than 60k of wild images. As we don't have access to such massive 3D scan data, the same position map data can be an option to create the ground truth dense landmark.

\section{Methodology}

As discussed in the above section, we don't have access to large 3D scan data. So, generating position maps similar to \cite{feng2018joint} can be an alternative. The position map records the 3D shape of the complete face in UV space as a 2D representation, where each pixel value has the 3D position information of that pixel. It provides correspondence to the semantic meaning of each point on the UV space. Their method aligns a 3D face model to the corresponding 2D face image and stores the 3D position of the points. We can apply the same to extract dense key points to create the ground truth data, before using direct regression to train a model that can predict those dense landmarks in 3D space. The following sections provide further details:

\subsection{Dense Facial Landmark Data Generation from UV Map}
As stated above,  \cite{feng2018joint} proposed a 3D facial representation based on the UV position map. They used the UV space to store the 3D position points from the 3D face model aligned with the 2D facial image. They assume the projection from the 3D model on the 2D image as a weak perspective projection and define the 3D facial position as a Left-hand Cartesian coordinate system. The ground truth 3D facial shape exactly matches the 2D image when projected to the x-y plane. They define the position map as \(Pos(u_i, v_i) = (x_i, y_i, z_i)\), where \((u_i, v_i)\) represents the \(i\)th point in face surface and \((x_i, y_i, z_i)\) represents the corresponding 3D position of facial mesh with \((x_i, y_i)\) being the corresponding 2D position of the face in the input RGB image and \(z_i\) representing the depth value of the corresponding point. 



\begin{figure}[t!] 
\centering    
\includegraphics[width=0.5\textwidth]{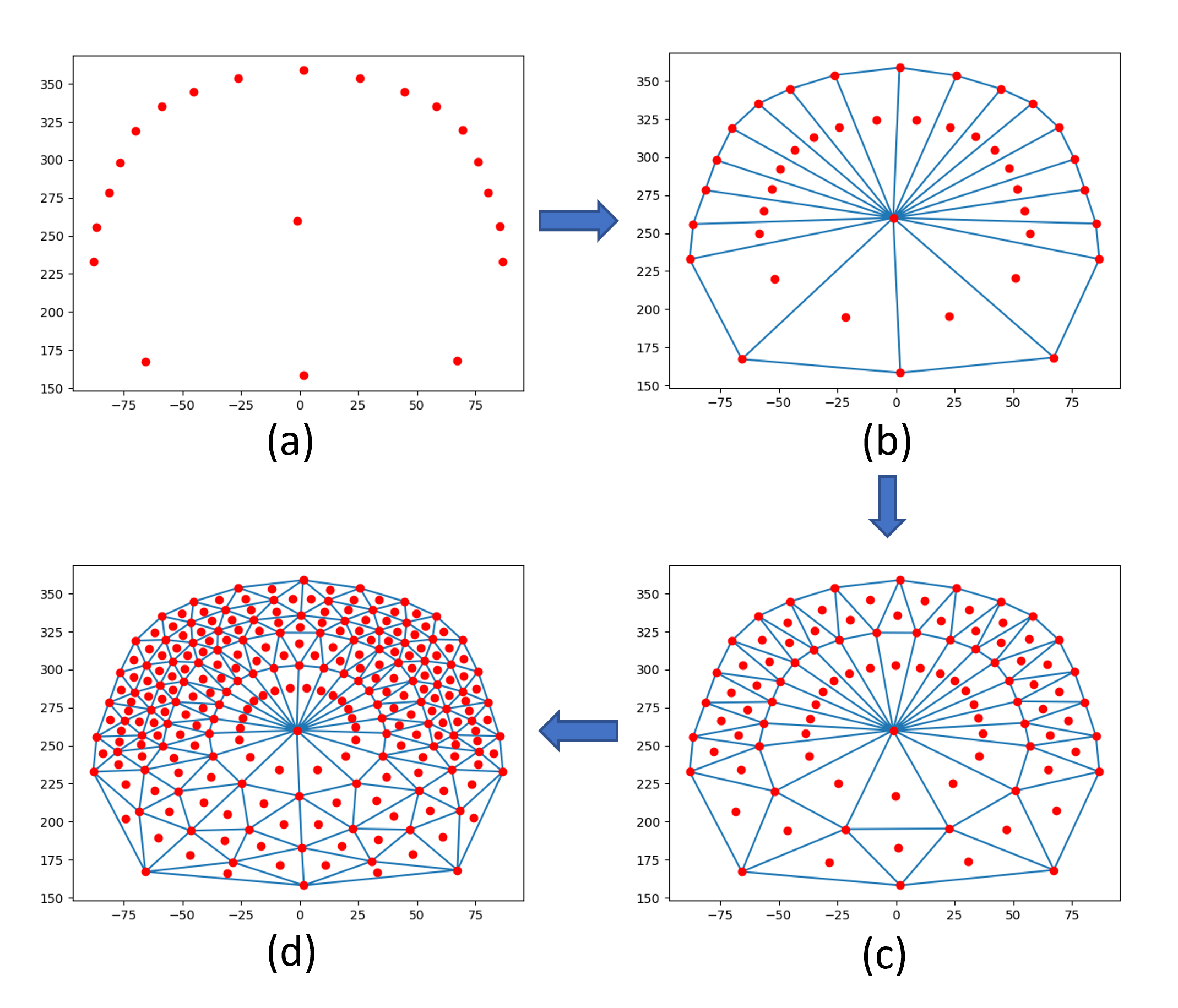}
\caption {Selection of key points through Delaunay Triangulation. (a) Initial selected key points across the jaw, forehead, and nose tip (b) First iteration of Delaunay triangulation and centroid selection (c) Second iteration of Delaunay triangulation and centroid selection (d) Third iteration of Delaunay triangulation and centroid selection}
\label{fig:delaunay_iterations}
\end{figure}

We followed the same representation and used their pipeline to build the raw data from the 300W-LP \cite{zhu2016face} dataset. This contains more than 60k unconstrained face images with fitted 3DMM parameters which are based on the Basel Face Model. They used the parameterized UV coordinates from \cite{bas20173d}, which compute a Tutte embedding \cite{floater1997parametrization} with conformal Laplacian weight and then map the mesh boundary to a square. So we can filter this UV position map data to create a dense face landmark. The 3DMM face template that was used by \cite{feng2018joint} has a total of 43867 vertices. Out of these, we have sampled 520 vertices. To sample, we have followed the following steps as shown in figure \ref{fig:delaunay_iterations} -
\begin{itemize}
    \item First, we have selected 18 key points across the jaw and one key point on the nose tip from the 68 key points provided.
    \item Then we run the Delaunay triangulation on the selected points and select the centroids of the three vertices of each triangle.
    \item We repeat the same step another two times and have the final key points.
    \item Finally, we select these key points across the template mesh and manually select the rest of the key points and rectify some of the already selected key points in Blender.
\end{itemize}


After these iterations, the final version of the ground truth data has the RGB face images and their corresponding 68 face key points and the selected 520 key points. The whole dataset contains around 61k pairs of ground truth images and their corresponding ground truth position map data saved in numpy format. Further, we expanded the data by applying a horizontal flip which made the total dataset size to 120k of paired images and their position map data.

\subsection{Dense Facial Landmark Prediction using Regression}

As we have around 120k pairs of ground truth face images in the wild and their corresponding ground truth 520 facial key points, we formulate the problem as a direct regression of those 520 key points from a monocular face image. We build a model with a standard feature extractor with a regressor head. The trained model will predict a continuous value of three positions (x,y,z) for those 520 3D landmarks. We choose the total number of classes as 520 x 3 = 1560. As the feature extractor, we have chosen two popular backbones, Resnet50 and MobilenetV2. 

The standard loss function that is typically used for any landmark estimator is the $L1$, and $L2$ loss or the Mean Square Error loss. But the $L2$ loss ($L2(x)\ =\ x^2/2$) function is sensitive to outliers, therefore \cite{rashid2017interspecies} used $smoothL1$ loss where they updated the $L2$ loss value with $|x|-1/2$ for $x>=1$.



Both $L1$ ($L1(x)\ =\ |x|$) and $smoothL1$ perform well for outliers, but they produce a very small value for small landmark differences. This hinders the network training for small errors. To solve this issue, \cite{feng2018wing} proposed a new loss called Wing loss which pays more attention to small and medium errors. They combined the $L1$ loss for the large landmark deviations and $ln(.)$ for small deviations as follows - 
\begin{equation}
    wing(x) = \left\{\begin{matrix}
w\ln(1+|x|/\epsilon), & if\ |x| < w\\ 
|x|-C, & otherwise
\end{matrix}\right.
\end{equation}\label{eq:wingLoss}

where $C\ =\ w-w\ln(1+w/\epsilon)$, $w$ and $\epsilon$ are the hyperparameters ($w$ = 15, $\epsilon$ = 3 in the paper). In this work as well we combined the Meas Square Error loss with the Wing loss to define a hybrid loss function as - 
\begin{equation}
    L = w_1L_{Wing} + w_2L_{MSE}
\end{equation}
Through an empirical study, we set the weight of these two loss terms as $w_1$ = 1.5 and $w_2$ = 0.5.

\section{Evaluation}

As we don't have any separate evaluation or test dataset that has the 3DMM parameters or the position map data available, we evaluated our trained model on the 3D face alignment task. To measure the face alignment quantitatively, we use the normalized mean error (NME) as the evaluation metric. NME is computed as the normalized mean Euclidean distance between each set of corresponding landmarks in the predicted result $l$ and the ground truth $l^{'}$ :
\begin{equation}\label{EQ14}
    \textup{NME} = \frac{1}{N}\sum_{i=1}^{N}\frac{\|l_i - l^{'}_i\|_2}{d}
\end{equation}

Following the previous works \cite{ruan2021sadrnet}, the normalization factor d is computed as $\sqrt{h \ast w}$, where $h$ and $w$ are the height and width of the bounding box, respectively. Similar to \cite{feng2018joint} and \cite{ruan2021sadrnet} for 2D and 3D sparse alignment, we consider all 68 landmark points. We divide the dataset based on the yaw angles $(0\degree, 30\degree)$, $(30\degree, 60\degree)$ and $(60\degree, 90\degree)$ and a balanced subset created by taking a random sample from the whole dataset. We benchmarked our model on the widely used AFLW2000-3D dataset. It is an in-the-wild face dataset with a large variation in illumination, pose, occlusion, and expression. It has 2000 images with 68 3D face landmark points for face alignment. The results are shown in table \ref{table:compareAFLW3D}.

\begin{table}[t!]
\caption{Quantitative evaluation on AFLW2000-3D dataset on facial alignment task Metrics - NME (Lower is better) for different Head Pose Bins}
\centering
\label{table:compareAFLW3D}
\begin{tabular}{|c|c|c|c|c|}
\hline 
Method & 0 to 30 & 30 to 60 & 60 to 90 & All  \\ 
\hline
3DDFA \cite{zhu2016face} & 3.43 & 4.24 & 7.17 & 4.94 \\
\hline
3DSTN \cite{bhagavatula2017faster} & 3.15 & 4.33 & 5.98 & 4.49 \\
\hline
3D-FAN \cite{bulat2017far} & 3.16 & 3.53 & 4.60 & 3.76 \\
\hline
3DDFA TPAMI \cite{zhu2017face} & 2.84 & 3.57 & 4.96 & 3.79 \\
\hline
PRNet \cite{feng2018joint} & 2.75 & 3.51 & 4.61 & 3.62 \\
\hline
2DASL \cite{tu20203d} & 2.75 & 3.46 & 4.45 & 3.55 \\
\hline
3DDFA V2\cite{guo2020towards} & 2.63 & 3.420 & 4.48 & 3.51 \\
\hline 
Ours & 2.86 & 3.68 & 4.76 & 3.77 \\
\hline 
\end{tabular}
\end{table}

\begin{table}[t!]
\caption{Quantitative evaluation on AFLW dataset with 21-point landmark definition on facial alignment task Metrics - NME (Lower is better) for different Head Pose Bins}
\centering
\label{table:compareAFLW}
\begin{tabular}{|c|c|c|c|c|}
\hline 
Method & 0 to 30 & 30 to 60 & 60 to 90 & All  \\ 
\hline
ESR \cite{cao2014face} & 5.66 & 7.12 & 11.94 & 8.24 \\
\hline
3DDFA \cite{zhu2016face} & 4.75 & 4.83 & 6.39 & 5.32 \\
\hline
3D-FAN \cite{bulat2017far} & 4.40 & 4.52 & 5.17 & 4.69 \\
\hline
3DDFA TPAMI \cite{zhu2017face} & 4.11 & 4.38 & 5.16 & 4.55 \\
\hline
PRNet \cite{feng2018joint} & 4.19 & 4.69 & 5.45 & 4.77 \\
\hline
3DDFA V2\cite{guo2020towards} & 3.98 & 4.31 & 4.99 & 4.43 \\
\hline 
Ours & 4.04 & 4.45 & 5.2 & 4.57 \\
\hline 
\end{tabular}
\end{table}

Following 3DDFA-V2 \cite{guo2020towards}, we have also evaluated our work using the AFLW full set (21K test images with 21-point landmarks). We followed the same split and showed the results for different angles in table \ref{table:compareAFLW}. 

\begin{table}[t!]
\caption{Comparative analysis with two different backbones Mobilenet-V2 and Resnet-18 of quantitative results on AFLW-3D dataset on facial alignment task and the computational requirement (gMac, gFlop, \# params - Number of parameters) Metrics - NME (Lower is better) for different Head Pose Bins}
\centering
\label{table:compareBackbone}
\begin{tabular}{|c|c|c|c|c|c|c|c|}
\hline 
Backbone & 0 to 30 & 30 to 60 & 60 to 90 & All & gMac & gFlop & \# Params   \\ 
\hline
Resnet-18 & 2.88 & 3.72 & 4.82 & 3.83 & 5.13 & 2.56 & 16.03M\\
\hline
Mobilenet-V2 & 2.86 & 3.68 & 4.76 & 3.77 & 0.39 & 0.19 & 4.18M\\
\hline
\end{tabular}
\end{table}

\begin{figure}[t!] 
\centering    
\includegraphics[width=0.8\textwidth]{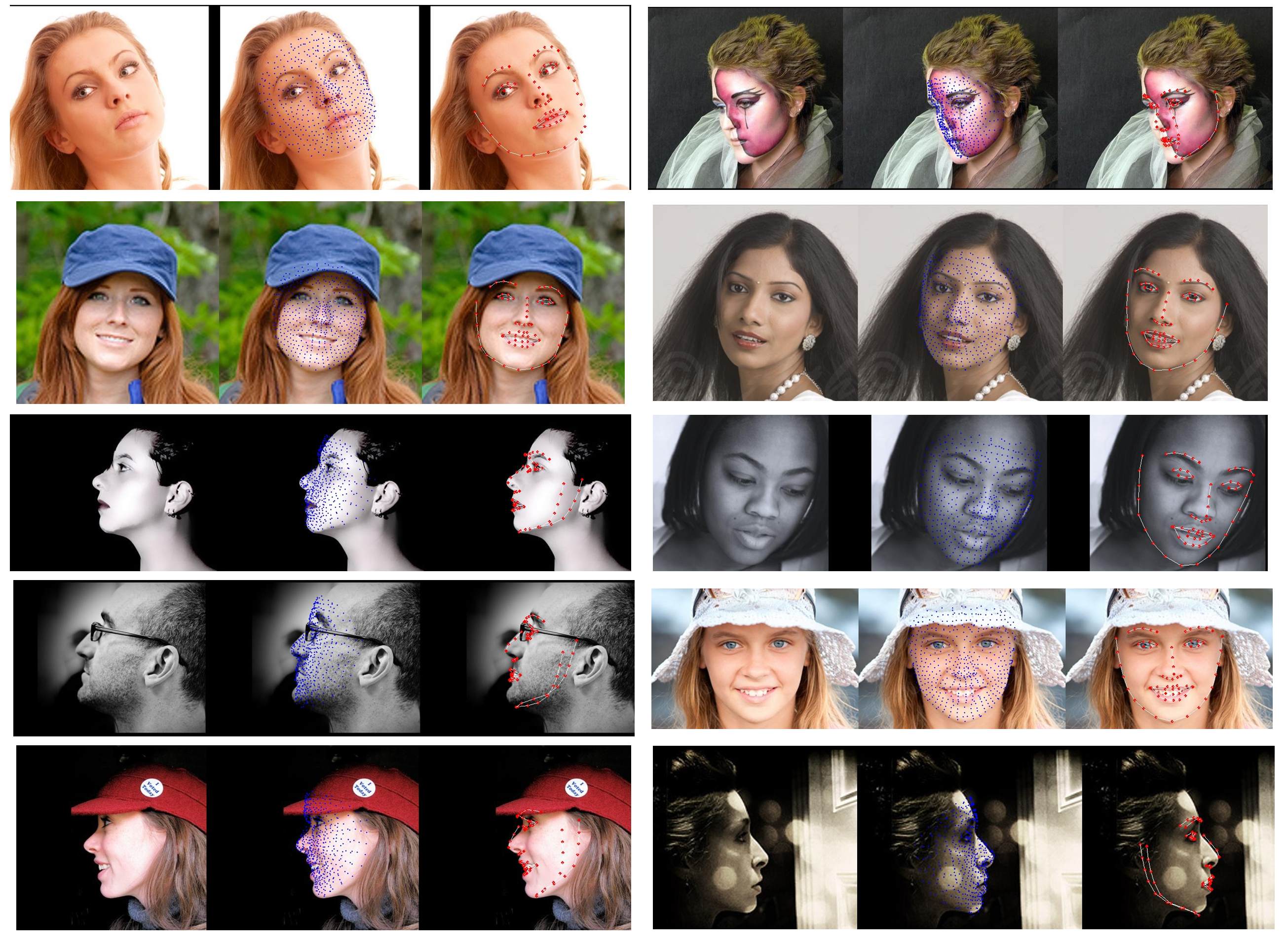}
\caption {Qualitative results: the first image shows the ground truth image, the second image shows the 520 key points and the third image shows the 68 key points predicted by the model}
\label{fig:results}
\end{figure}

\begin{figure}[!tbp]
  \centering
  \subfloat[]{\includegraphics[width=0.4\textwidth]{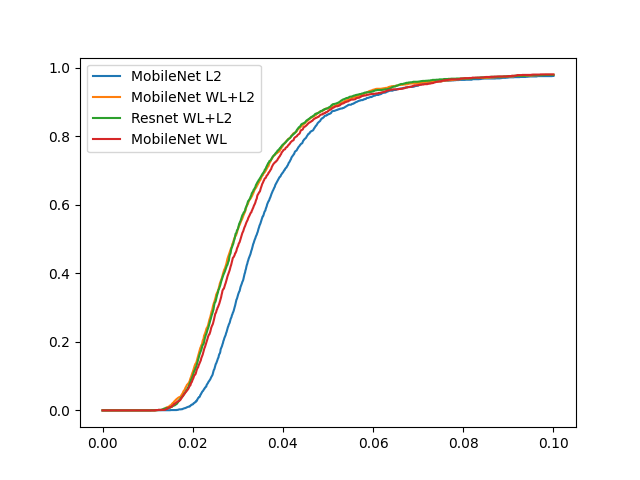}\label{fig:f1}}
  \hfill
  \subfloat[]{\includegraphics[width=0.4\textwidth]{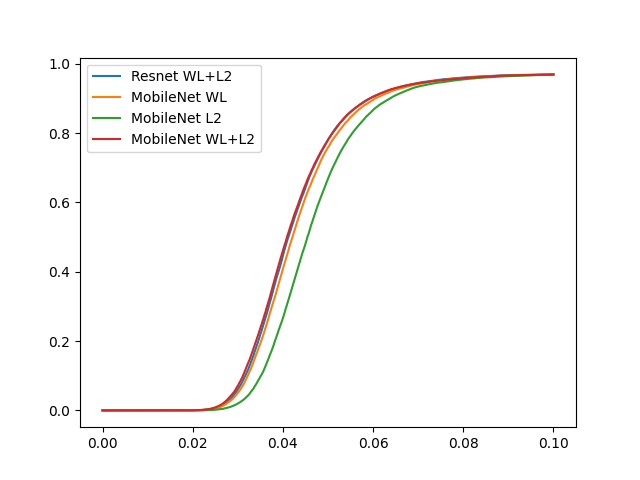}\label{fig:f2}}
  \caption{Different loss function study on Cumulative Errors (NME) Distribution (CED) curves on (a) AFLW2000-3D with 68 point landmarks and (b) AFLW with 21 point landmarks. The backbone and loss functions are also shown in the legend. WL stands for Wing Loss, and L2 stands for MSE loss.}
  \label{fig:AUC_AFLW}
\end{figure}

\section{Discussion}
As we do not have access to any dense landmark evaluation dataset, we evaluated our trained network against the 3D face alignment task for 68 points FLD on AFLW2000 and for 21 points FLD on AFLW dataset. The experimental results in table \ref{table:compareAFLW3D} and \ref{table:compareAFLW} show that our model is able to outperform most of the previous methods. Also as we have used the MobilenetV2-based model, the overall model size is comparatively small and requires less amount of computational resources. Table \ref{table:compareBackbone} shows a comparative analysis of the Rensent and Mobilenet backbone-based networks in terms of their computational resource requirement. We have also conducted a study on the effect of the hybrid loss function. Figure \ref{fig:AUC_AFLW} shows the cumulative error distribution curves based on NME for the AFLW-3D and AFLW datasets. In both cases, a combination of Wing Loss and MSE performs better than the rest. Figure \ref{fig:results} shows some of the qualitative results on some samples from the AFLW dataset. The model shows a comparative result on extreme head pose angles as well as with occlusions and other facial variations.

\section{Conclusion}

In this work, we have presented a methodology to create a dense landmark dataset that has 520 key points generated from the UV position map data. With the help of generated dataset, we have trained an FLD regressor network with two different backbones, Resnet18 and MobileNetV2. As we do not have access to any other dense landmark evaluation dataset we have evaluated our trained model on a 68 points FLD task. Experimental results show our trained model is able to outperform most of the existing landmark detection methods while using fewer computational resources. The qualitative results show the robustness of our model and provide better results in extreme head pose angles. Though visually, the results of the model look good, due to the lack of ground truth test data, we are only able to evaluate the model against the 3D facial alignment task. In the future, we can extend this work and use those 520 key points to fit an existing statistical (e.g., 3DMM) model to the face and evaluate the full face reconstruction benchmark.

\section*{Acknowledgments}
This work was supported by The Science Foundation Ireland  Centre for Research Training in Digitally-Enhanced Reality(d-real) under Grant No.18/CRT/6224. Some of this work is done as a part of an Internship at Xperi Corporation.


\bibliographystyle{apalike}

\bibliography{imvip}

\end{document}